\documentclass[10pt,twocolumn,letterpaper]{article}

\usepackage{cvpr}              

\usepackage{placeins}

\definecolor{cvprblue}{rgb}{0.21,0.49,0.74}
\usepackage[pagebackref,breaklinks,colorlinks,allcolors=cvprblue]{hyperref}

\def\paperID{72} 
\def\confName{CVPR}
\def\confYear{2025}

\title{Enhancing Hazy Wildlife Imagery: AnimalHaze3k and IncepDehazeGan}

\author{Shivarth Rai\\
{\tt\small raishivarth@gmail.com}
\and
Tejeswar Pokuri\\
{\tt\small tejeswarpokuri3@gmail.com}
\and 
\\ Department of Computer Science\\
Manipal Institute of Technology, Manipal, India
}
\begin{document}

\maketitle
\begin{abstract}
Atmospheric haze significantly degrades wildlife imagery, impeding computer vision applications critical for conservation, such as animal detection, tracking, and behavior analysis. To address this challenge, we introduce AnimalHaze3k–a synthetic dataset comprising of 3,477 hazy images generated from 1,159 clear wildlife photographs through a physics-based pipeline. Our novel IncepDehazeGan architecture combines inception blocks with residual skip connections in a GAN framework, achieving state-of-the-art performance (SSIM: 0.8914, PSNR: 20.54, and LPIPS: 0.1104),  delivering 6.27\% higher SSIM and 10.2\% better PSNR than competing approaches. When applied to downstream detection tasks, dehazed images improved YOLOv11 detection mAP by 112\% and IoU by 67\%. These advances can provide ecologists with reliable tools for population monitoring and surveillance in challenging environmental conditions, demonstrating significant potential for enhancing wildlife conservation efforts through robust visual analytics. The AnimalHaze3k dataset can be publicly accessed at: \url{https://shvrth.github.io/}.

\end{abstract}
\section{Introduction}
\label{paper_sec:introduction}

Computer vision-based methods have become indispensable tools for wildlife conservation and ecological research, enabling non-invasive monitoring of animals, population tracking, and behavioral studies across diverse habitats \cite{hou2020identification}\cite{schindler2021identification}. These methods have become critical for conservation strategies, such as identifying animals through unique markings \cite{chalmers2019conservation}, detecting elusive and endangered species \cite{ren2021tracking}\cite{guan2023face}, and monitoring illegal poaching activities of protected animal species \cite{chalmers2019conservation}. Their ability to process large volumes of data with high accuracy and consistency \cite{norouzzadeh2018automatically}, and provide cost-effective solutions to real-time animal monitoring have made their use widespread \cite{saoud2025hubot}\cite{o2011camera}. However, environmental challenges like atmospheric haze greatly compromise image quality, undermining the effectiveness of these technologies \cite{newey2015limitations}.

Haze-induced degradation manifests through unnatural color shifts, decreased visibility, diminished contrast between subjects and backgrounds, and distorted depth perception \cite{wang2017recent}\cite{singh2019comprehensive}. Such quality deterioration directly impacts downstream conservation tasks—diminished visibility impedes movement tracking, low-contrast images complicate animal detection and population counting, and can obscure anatomical features challenging automation models for species identification. 

Image dehazing aims to generate the latent haze-free image from the observed hazy image. The atmospheric scattering model \cite{narasimhan2002vision}\cite{narasimhan2003interactive} is the classical framework for modeling hazy image formation through light-particle interactions. This model describes hazy images as a per-pixel combination of attenuated scene radiance and atmospheric light interference:
\begin{equation}
  I(x) = J(x)\cdot t(x) + A\cdot(1- t(x))
  \label{eq:asm}
\end{equation}
where $x$ is the pixel index, $I(x)$ is observed hazy image, $J(x)$ is the clean scene radiance,  $A$ denotes the global atmospheric light, and $t(x)$ is the transmission matrix defined as:
\begin{equation}
	t(x) = e^{ - \beta \cdot d(x)}
  \label{eq:tx}
\end{equation}
where $\beta$ represents the scattering coefficient of the atmosphere, and $d(x)$ denotes the depth information.

Recently, single image dehazing has made significant progress with the use of data-driven approaches utilizing synthetic image pairs to achieve superior performance. Initial CNN-based architectures \cite{cai2016dehazenet}\cite{ren2016single}\cite{zhang2018densely} employed a decomposed estimation framework, separately predicting atmospheric light $A$ and transmission maps $t(x)$, with supervision derived from synthetic transmission ground truth. Modern approaches \cite{dong2020multi}\cite{dong2020physics}\cite{wu2021contrastive} have pivoted toward direct prediction of latent haze-free images or their residual components relative to hazy inputs, capitalizing on pixel-level reconstruction to optimize perceptual quality. ViT-based methods \cite{vaswani2017attention}\cite{song2023vision} have outperformed CNN-based models through their use of global attention mechanisms.

\begin{figure}[ht]
  \centering
  \captionsetup{justification=centering}
  \begin{subfigure}[b]{0.32\linewidth}
    \includegraphics[width=\linewidth]{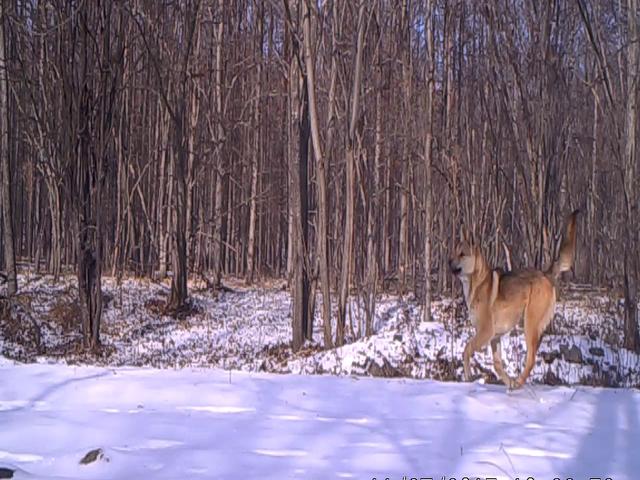}
    \caption{}
    \label{fig:row1-1}
  \end{subfigure}
  \hfill
  \begin{subfigure}[b]{0.32\linewidth}
    \includegraphics[width=\linewidth]{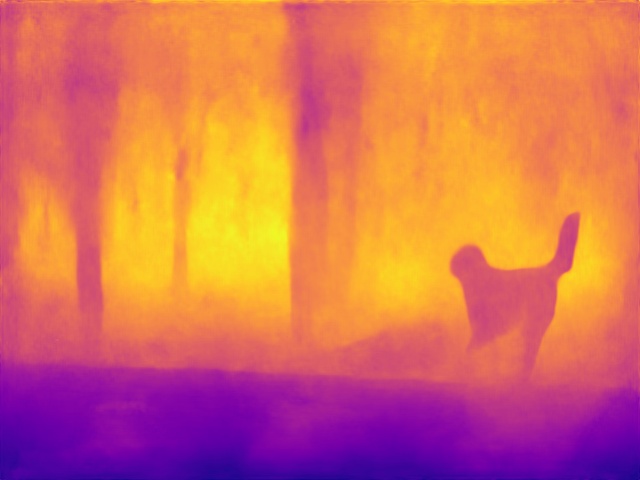}
    \caption{}
    \label{fig:row1-2}
  \end{subfigure}
  \hfill
  \begin{subfigure}[b]{0.32\linewidth}
    \includegraphics[width=\linewidth]{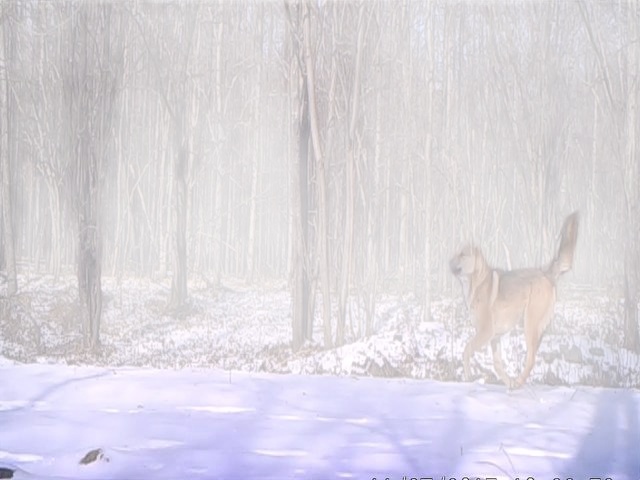}
    \caption{}
    \label{fig:row1-3}
  \end{subfigure}
  \hfill
 
  \vspace{0.4cm} 
  \begin{subfigure}[b]{0.32\linewidth}
    \includegraphics[width=\linewidth]{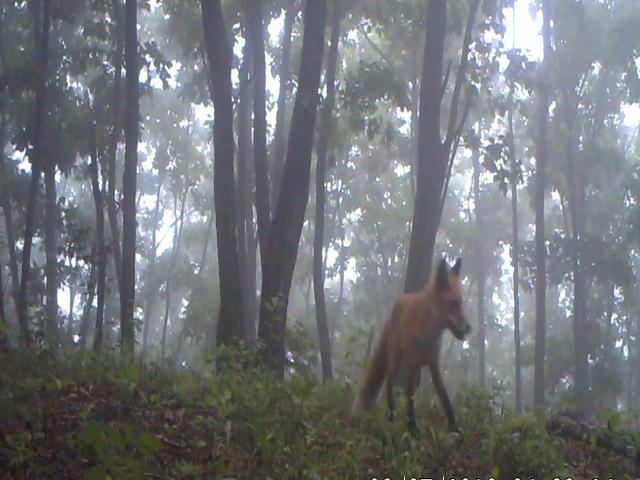}
    \caption{}
    \label{fig:row2-1}
  \end{subfigure}
  \hfill
  \begin{subfigure}[b]{0.32\linewidth}
    \includegraphics[width=\linewidth]{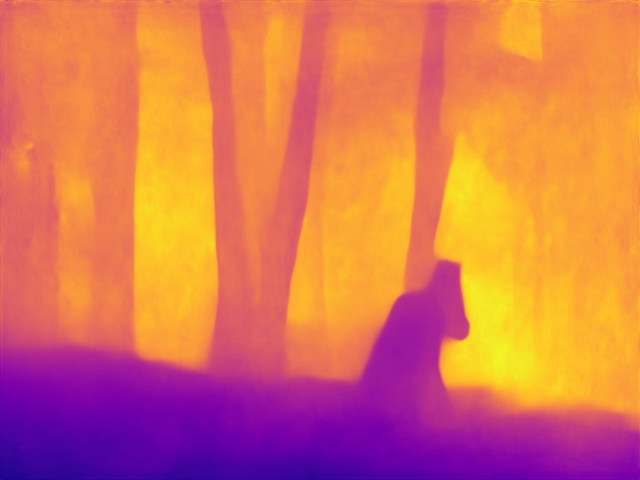}
    \caption{}
    \label{fig:row2-2}
  \end{subfigure}
  \hfill
  \begin{subfigure}[b]{0.32\linewidth}
    \includegraphics[width=\linewidth]{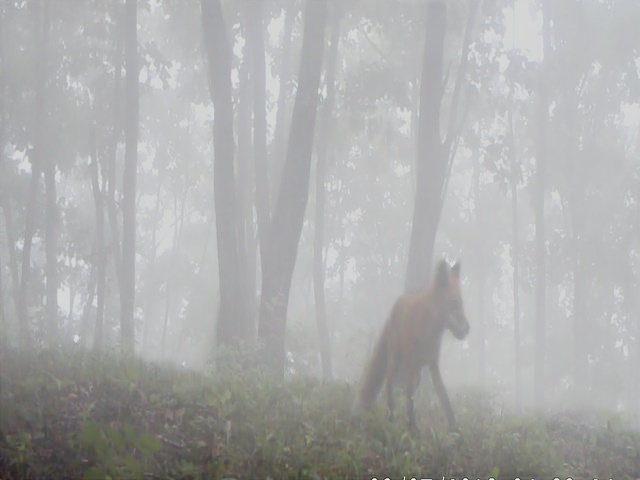}
    \caption{}
    \label{fig:row2-3}
  \end{subfigure}
  \hfill

  \caption{Samples from AnimalHaze3k dataset showing the ground truth, depth map used for synthetic haze generation and hazy image generated for a (a-c) Dog, and, (d-f) Red Fox.}
  \label{fig:data sample}
\end{figure}

While recent methodological advancements demonstrate considerable promise, their efficacy remains constrained by a critical bottleneck: the scarcity of paired real-world training data exhibiting authentic haze conditions. For the task of dehazing wildlife imagery, acquiring such datasets poses significant challenges due to environmental variability, and the inherent difficulty of capturing the same haze-free and hazy scene in dynamic wildlife habitats. In response to these data acquisition challenges, contemporary dehazing approaches synthesize training pairs by artificially inducing haze in clean images. This process involves estimating scene depth—either leveraging existing depth maps from specialized datasets or deriving them algorithmically—and simulating haze formation through the atmospheric scattering model using \cref{eq:asm}. 

This work aims to overcome the aforementioned hurdles and makes two-fold technical contributions:
\begin{itemize}
    \item AnimalHaze3k dataset: A synthetic dataset of 3,477 hazy wildlife images generated from 1,159 real wildlife photographs via a physics-based pipeline, simulating diverse atmospheric haze conditions.
    \item IncepDehazeGan: A GAN architecture integrating inception blocks, residual skip connections, and a hybrid adversarial-L1 loss, achieving state-of-the-art dehazing on the proposed AnimalHaze3k dataset.
\end{itemize}

\section{The AnimalHaze3k Dataset}
\label{paper_sec:dataset}

The dataset used for acquiring clear images to generate synthetic hazy images is the Northeast Tiger and Leopard National Park (NTLNP) dataset\cite{tan2022animal}. This dataset contains 25,567 images of 17 species captured under both daylight and night-time conditions. Only daylight images were utilized in this study. The images were obtained through infrared camera traps deployed in China's Northeast Tiger and Leopard National Park between 2014 and 2020. The camera trap methodology ensures animals were observed in their natural, undisturbed habitats. The dataset features diverse environmental contexts including varied backgrounds, weather conditions, and seasonal variations for non-hibernating species. From this collection, we systematically selected 1,159 images representing 11 species. All images were pre-processed to remove temporal stamps (date and time) and were standardized to $640\times480$ pixel resolution.

\begin{figure}[ht]
  \centering
  \captionsetup{justification=centering}
  \includegraphics[width=1\linewidth]{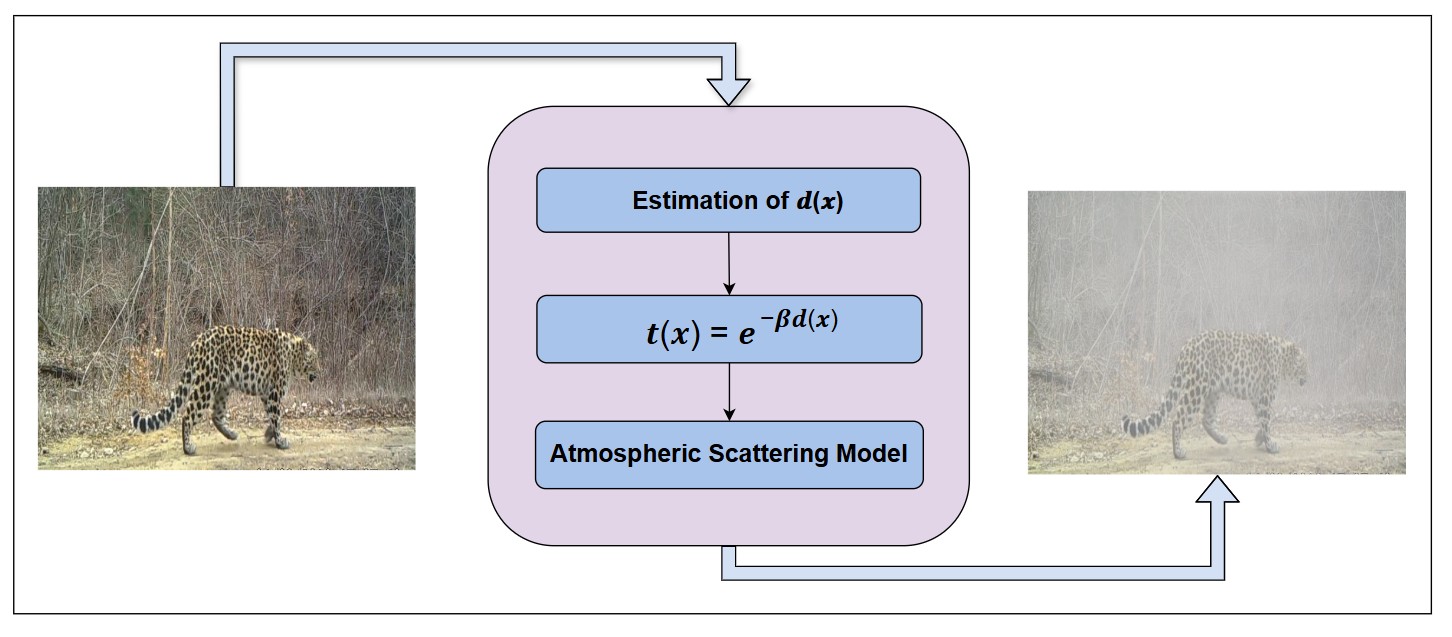} 
  \caption{Data generation pipeline stages: (1) Depth estimation, (2) Transmission map calculation, (3) Synthetic haze generation.}
  \label{fig:pipeline}
\end{figure}

The AnimalHaze3k data generation pipeline comprises three principal stages as illustrated in \cref{fig:pipeline}. In the first stage, a depth map for the input image is estimated using the HybridDepth\cite{ganj2024hybriddepth} model. The reasons to choose HybridDepth are threefold: (1) HybridDepth is a state-of-art single image metric depth estimation pipeline. By fusing depth-from-focus (DFF) data with relative depth priors, HybridDepth achieves superior metric accuracy and generalization over models like ZoeDepth\cite{bhat2023zoedepth}, DFV\cite{yang2022deep} and Depth Anything\cite{yang2024depth}. (2) HybridDepth maintains consistent depth estimations across different zoom levels - while Depth Anything and DFV can overestimate depth. (3) HybridDepth uses Depth Anything\cite{yang2024depth} as the relative depth estimator. Depth Anything has been trained on a vast amount of outdoor data that results in high quality synthetic hazy images generated.
In the second stage of the data generation pipeline, transmission map, $t(x)$, is calculated using \cref{eq:tx}. Value of scattering coefficient $\beta$ is uniformly sampled from [1.8, 3.0], in order to generate an arbitrary haze density for each input image and improve the diversity of the dataset. In the final stage, the synthetic hazy image is generated using Eqn.(2). Atmospheric ambient light, $A$, values are taken at random from discrete values [0.8, 0.85, 0.9, 0.95, 1] following established methodologies\cite{li2018benchmarking}. 

Each clear image generates three synthetic hazy variations through this pipeline, ensuring comprehensive coverage of potential atmospheric conditions. The training set contains 1,041 clear and 3,123 hazy images. The validation and test sets contain 59 clear and 177 hazy images each. The dataset thus contains paired clear and hazy images. Examples from the dataset are depicted in \cref{fig:data sample}.


\begin{figure*}[ht] 
  \centering
  \includegraphics[width=0.9\linewidth]{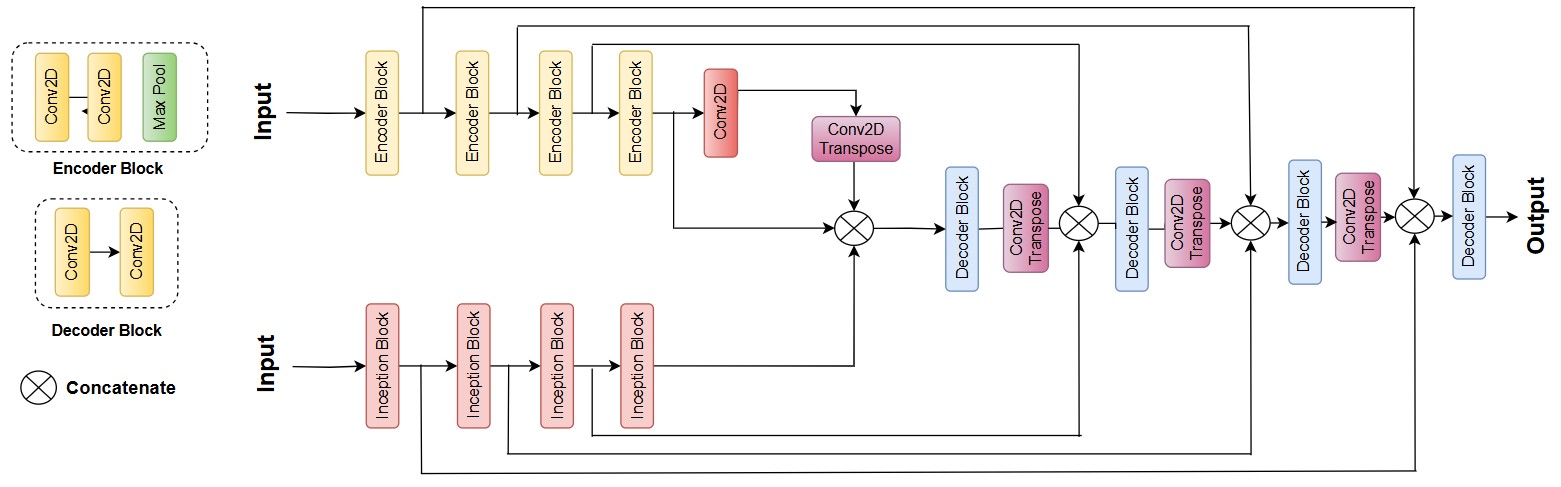}
  \caption{IncepDehazeGan Generator architecture}
  \label{fig:generator}
\end{figure*}

\section{The IncepDehazeGan Model}
\label{paper_sec:model}

IncepDehazeGan is a novel single-image dehazing GAN architecture. The IncepDehazeGan generator is a dense encoder-decoder network incorporating Inception Blocks\cite{inception}. The encoder consists of two parallel processing sections: (1) a series of standard encoding blocks made up of two convolution layers followed by ReLU activation function and a max pooling layer, and (2) a series of Inception Blocks using four parallel convolutional layers of distinct kernel sizes $(1\times1, 1\times3, 3\times1, 3\times3)$. 

\begin{figure}[!ht]
  \centering
  \captionsetup{justification=centering}
  \includegraphics[width=1\linewidth]{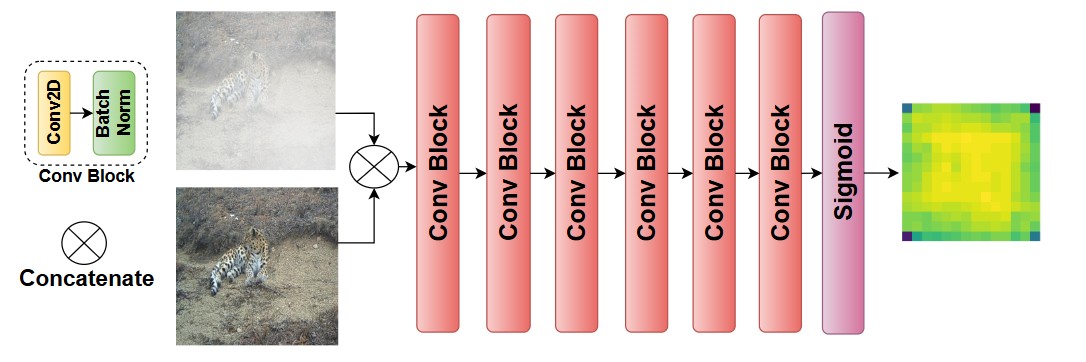} 
  \caption{IncepDehazeGan Discriminator architecture}
  \label{fig:discriminator}
\end{figure}

\begin{figure*}[ht]
  \centering
  \begin{subfigure}[b]{0.135\linewidth}
    \includegraphics[width=\linewidth]{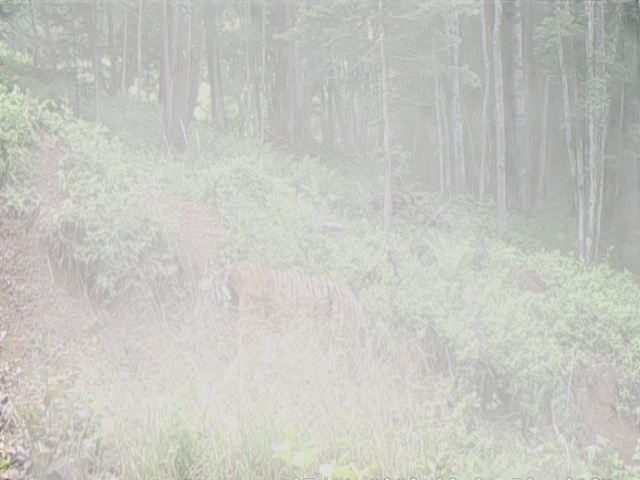}
    \caption{Hazy}
    \label{fig:tiger_hazy}
  \end{subfigure}
  \hfill
  \begin{subfigure}[b]{0.135\linewidth}
    \includegraphics[width=\linewidth]{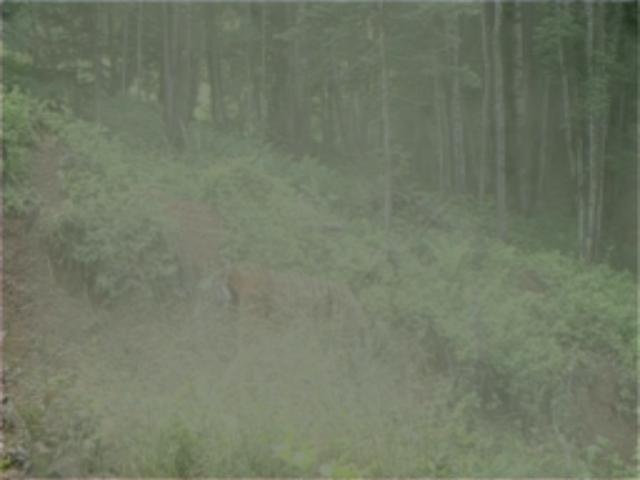}
    \caption{FFA}
    \label{fig:tiger_ffa}
  \end{subfigure}
  \hfill
  \begin{subfigure}[b]{0.135\linewidth}
    \includegraphics[width=\linewidth]{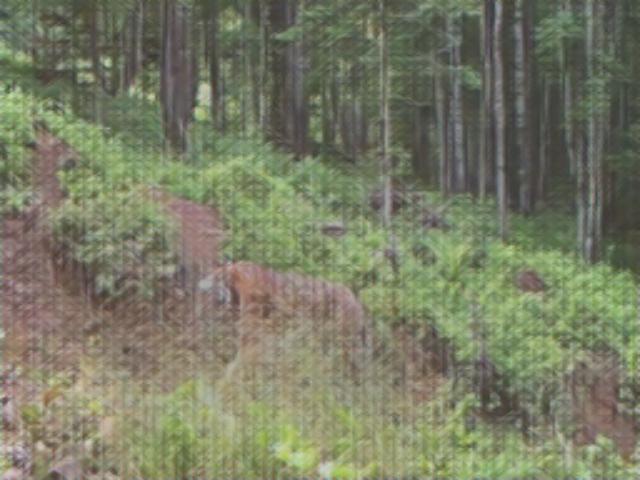}
    \caption{FD-GAN}
    \label{fig:tiger_fdgan}
  \end{subfigure}
  \hfill
  \begin{subfigure}[b]{0.135\linewidth}
    \includegraphics[width=\linewidth]{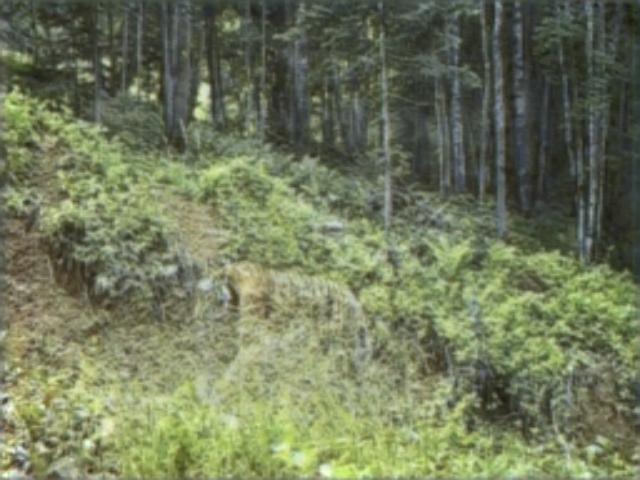}
    \caption{DEANET}
    \label{fig:tiger_deanet}
  \end{subfigure}
  \hfill
  \begin{subfigure}[b]{0.135\linewidth}
    \includegraphics[width=\linewidth]{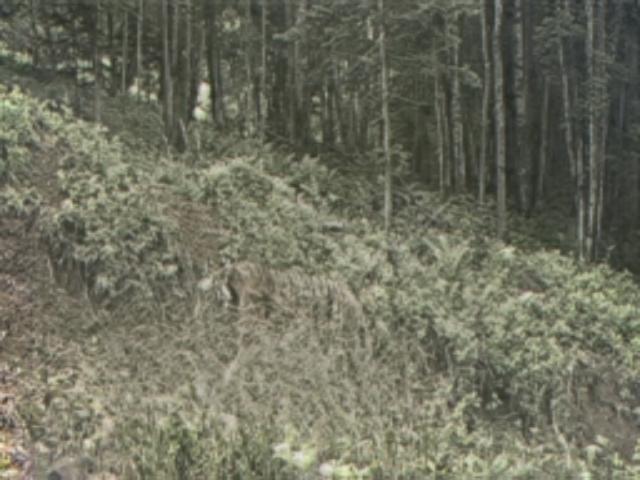}
    \caption{DehazeFormer}
    \label{fig:tiger_dehazeformer}
  \end{subfigure}
  \hfill
  \begin{subfigure}[b]{0.135\linewidth}
    \includegraphics[width=\linewidth]{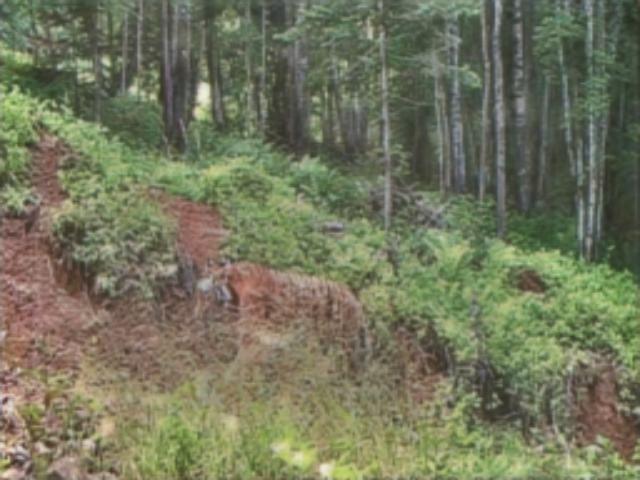}
    \caption{IncepDehaze}
    \label{fig:tiger_incep}
  \end{subfigure}
  \hfill
  \begin{subfigure}[b]{0.135\linewidth}
    \includegraphics[width=\linewidth]{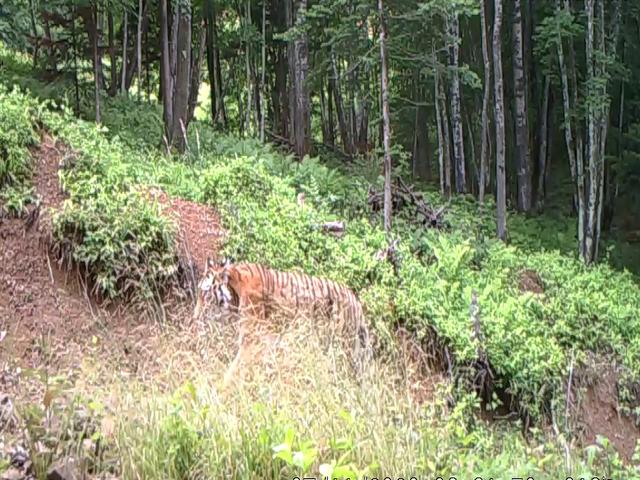}
    \caption{Ground Truth}
    \label{fig:tiger_clear}
  \end{subfigure}
  
  \vspace{0.5cm}
  
  \begin{subfigure}[b]{0.135\linewidth}
    \includegraphics[width=\linewidth]{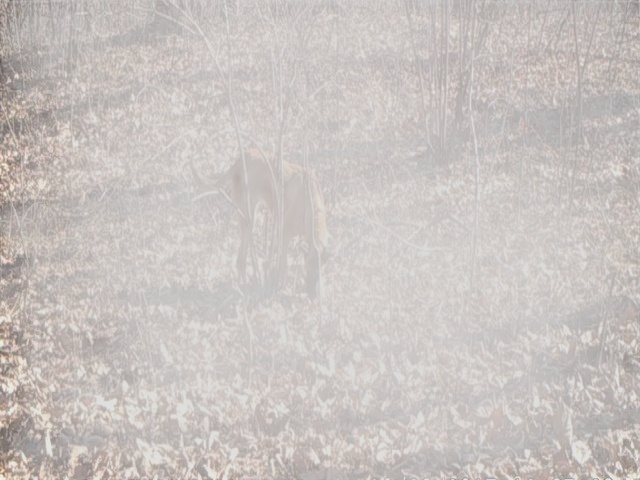}
    \caption{Hazy}
    \label{fig:dog_hazy}
  \end{subfigure}
  \hfill
  \begin{subfigure}[b]{0.135\linewidth}
    \includegraphics[width=\linewidth]{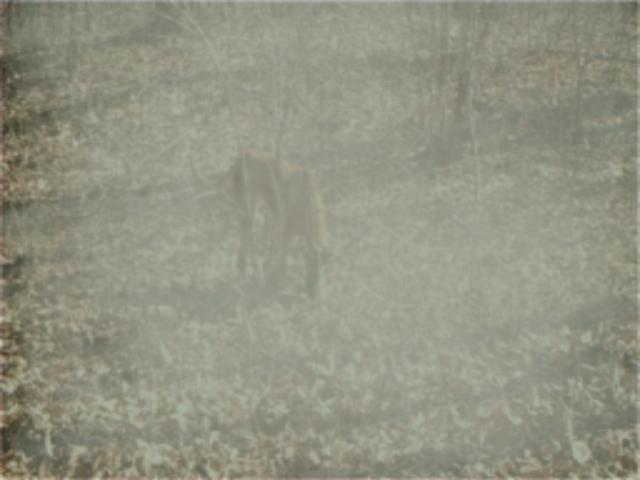}
    \caption{FFA}
    \label{fig:dog_ffa}
  \end{subfigure}
  \hfill
  \begin{subfigure}[b]{0.135\linewidth}
    \includegraphics[width=\linewidth]{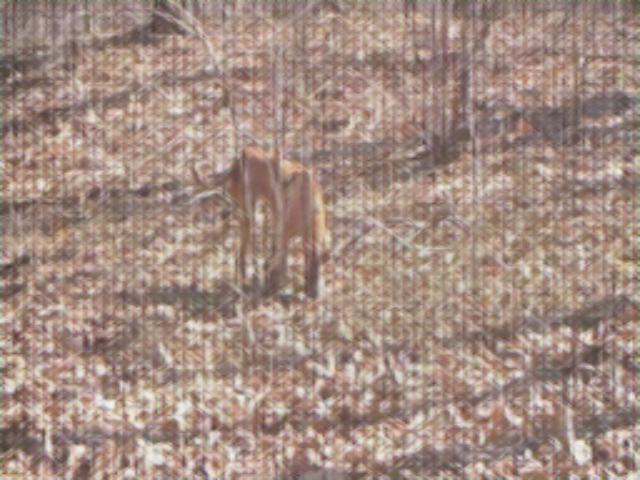}
    \caption{FD-GAN}
    \label{fig:dog_fdgan}
  \end{subfigure}
  \hfill
  \begin{subfigure}[b]{0.135\linewidth}
    \includegraphics[width=\linewidth]{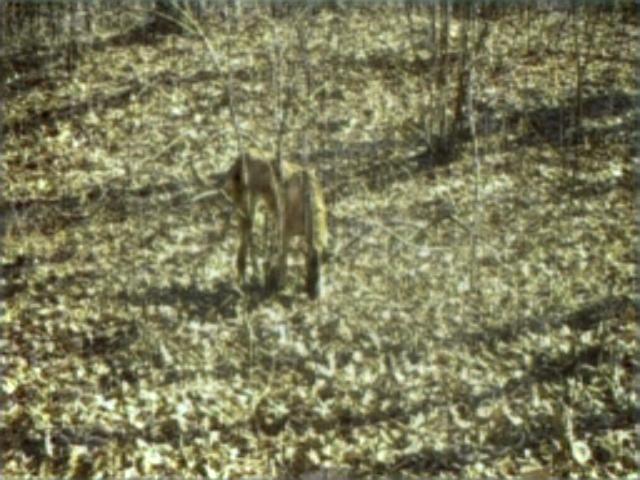}
    \caption{DEANET}
    \label{fig:dog_deanet}
  \end{subfigure}
  \hfill
  \begin{subfigure}[b]{0.135\linewidth}
    \includegraphics[width=\linewidth]{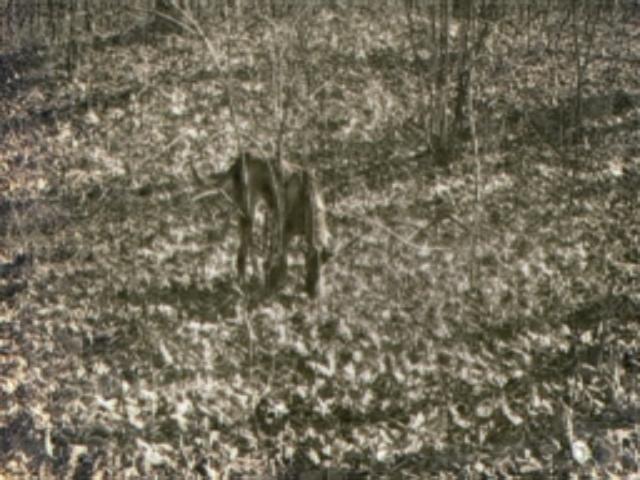}
    \caption{DehazeFormer}
    \label{fig:dog_dehazeformer}
  \end{subfigure}
  \hfill
  \begin{subfigure}[b]{0.135\linewidth}
    \includegraphics[width=\linewidth]{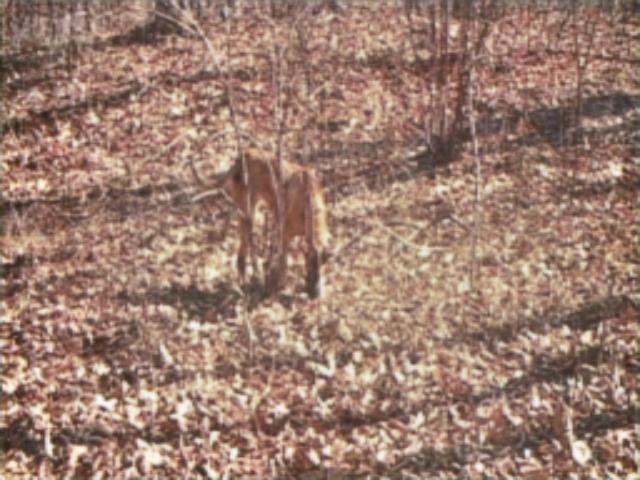}
    \caption{IncepDehaze}
    \label{fig:dog_incep}
  \end{subfigure}
  \hfill
  \begin{subfigure}[b]{0.135\linewidth}
    \includegraphics[width=\linewidth]{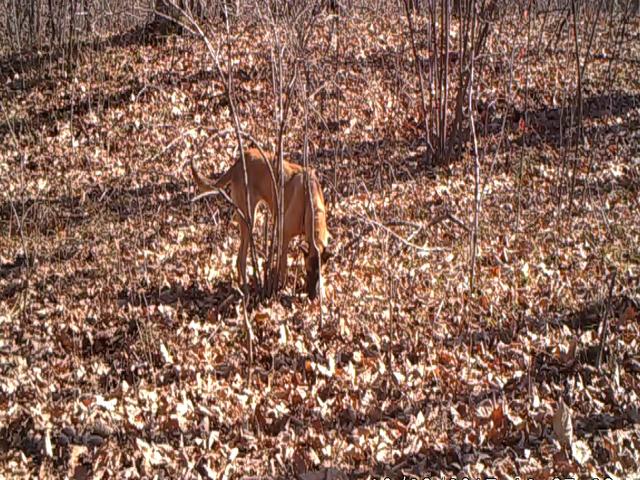}
    \caption{Ground Truth}
    \label{fig:dog_clear}
  \end{subfigure}
  
  \caption{Qualitative comparison on dehazing results across SOTA models and IncepDehazeGan (our model).}
  \label{fig:dehazing_comparison}
\end{figure*}

The decoder is a single series of blocks consisting of a transpose convolution layer preceded by two standard convolution layers. We use residual skip connections to share low-level feature maps learned in each encoder block with decoder blocks\cite{residuals}. This multilayer feature fusion reduces information loss due to down sampling operations in the encoder section\cite{multilayer}. The overall generator architecture is illustrated by \cref{fig:generator}. For training the generator, we use a combination of adversarial loss and mean absolute error. The adversarial loss is computed using the output of the discriminator and the target labels of ones, encouraging the generator to produce realistic outputs that deceive the discriminator. This loss is defined as follows:
\begin{equation}
\begin{aligned}
\mathcal{L}_{\text{Adv}}(x, y) ={} & -\frac{1}{N} \sum_{i=1}^{N} \Big[ y_i \log\left(\sigma(x_i)\right) \\
& + (1 - y_i) \log\left(1 - \sigma(x_i)\right) \Big]
\end{aligned}
\end{equation}

Mean absolute error ($L1$ loss) is calculated between the generated image and the target image. This loss term helps to maintain the fidelity of the generated output relative to the ground truth and is defined as: 
\begin{equation}
\mathcal{L}_{\text{L1}} = \frac{1}{N} \sum_{i=1}^{N} | y_i - \hat{y}_i |
\end{equation}

The net generator loss is a weighted sum of the adversarial loss and the L1 loss, where \(\lambda\) is a hyperparameter controlling the trade-off between these two losses. In our implementation, \(\lambda\) is set to 100. The net generator loss is thus expressed as:
\begin{equation}
\mathcal{L}_{\text{G}} = \mathcal{L}_{\text{Adv}} + \lambda \cdot \mathcal{L}_{\text{L1}}
\end{equation}

The IncepDehazeGan discriminator consists of six convolutional blocks followed by a final sigmoid activation. Each convolution block contains a convolutional layer with leaky ReLU activation and a batch normalization layer. \cref{fig:discriminator} depicts the IncepDehazeGan discriminator architecture. This produces a single-channel matrix, encouraging the generator to produce realistic local features and textures across the image\cite{pix2pix}. By focusing on small regions for each pixel, the discriminator also helps mitigate overfitting. For training the discriminator, we use Binary Cross Entropy to distinguish between ground truth and output produced by the generator. 

\section{Experimental Study}
\label{paper_sec:experiments}

Experiments in this study were conducted on a NVIDIA Tesla P100 GPU, with 16GB memory and 3584 CUDA cores. 

\subsection{Single-Image Dehazing}
We benchmark our AnimalHaze3k dataset using several state-of-the-art image dehazing models, FD-GAN\cite{DBLP:journals/corr/abs-2001-06968}, FFA-Net\cite{FFANET}, DehazeFormer\cite{song2023vision}, DEA-Net\cite{DEANET} and the proposed IncepDehazeGan model. The models were trained for 50 epochs, with batch size of 4. We extensively evaluated performance using several metrics-SSIM\cite{ssim}, PSNR\cite{psnr}, FSIM\cite{fsim} and LPIPS\cite{lpips}.

\begin{figure}[!ht]
  \centering
  \captionsetup{justification=centering}
  \begin{subfigure}[b]{0.32\linewidth}
    \includegraphics[width=\linewidth]{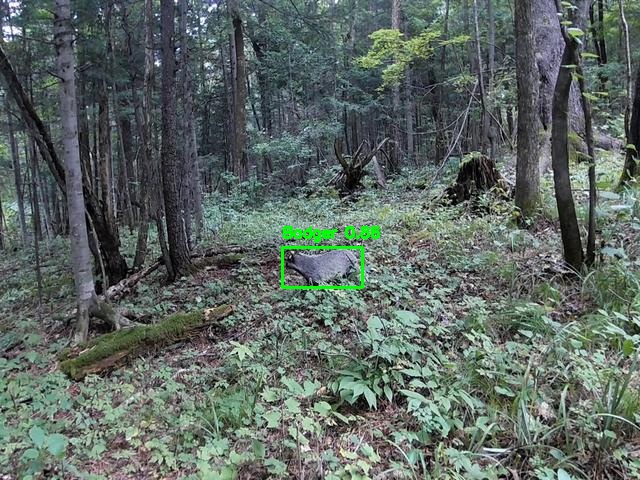}
    \caption{}
    \label{fig:triple_a}
  \end{subfigure}
  \hfill
  \begin{subfigure}[b]{0.32\linewidth}
    \includegraphics[width=\linewidth]{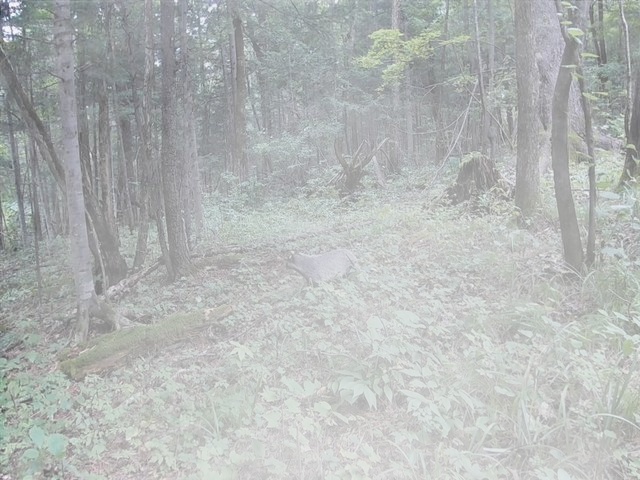}
    \caption{}
    \label{fig:triple_b}
  \end{subfigure}
  \hfill
  \begin{subfigure}[b]{0.32\linewidth}
    \includegraphics[width=\linewidth]{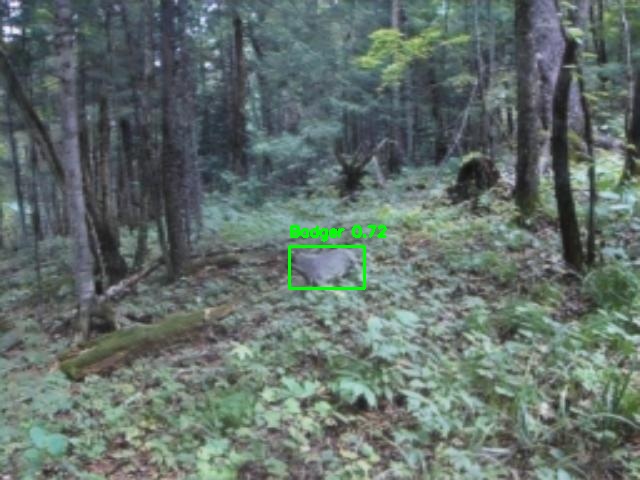}
    \caption{}
    \label{fig:triple_c}
    
  \end{subfigure}
  \begin{subfigure}[b]{0.32\linewidth}
    \includegraphics[width=\linewidth]{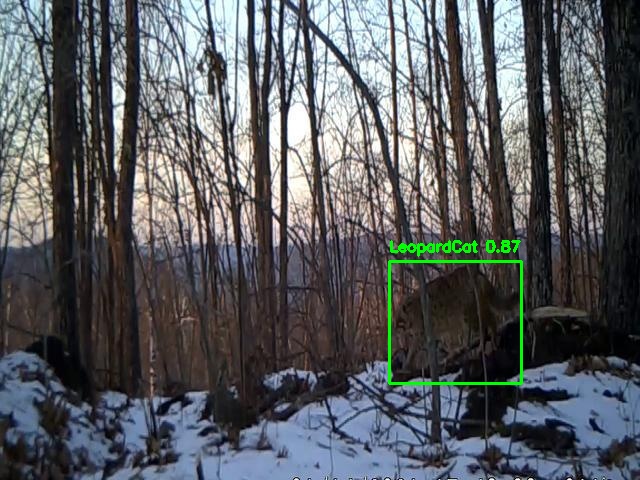}
    \caption{}
    \label{fig:triple_a}
  \end{subfigure}
  \hfill
  \begin{subfigure}[b]{0.32\linewidth}
    \includegraphics[width=\linewidth]{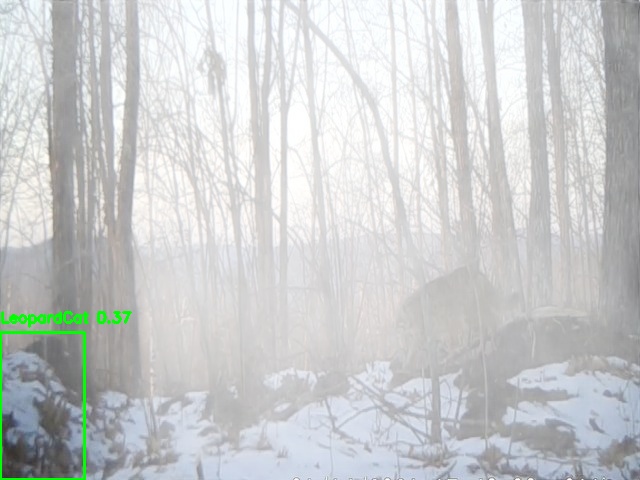}
    \caption{}
    \label{fig:triple_b}
  \end{subfigure}
  \hfill
  \begin{subfigure}[b]{0.32\linewidth}
    \includegraphics[width=\linewidth]{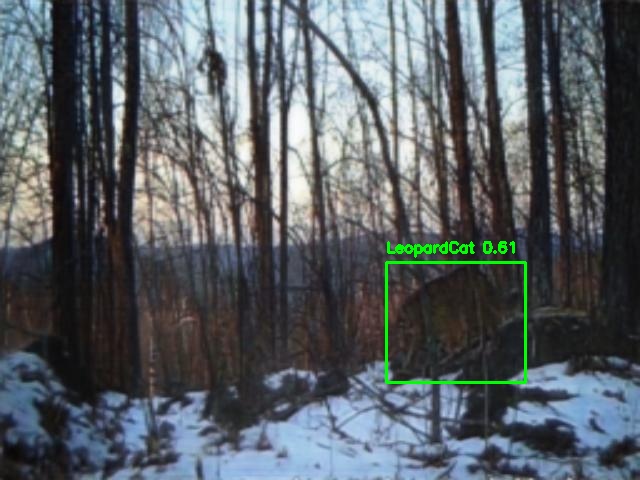}
    \caption{}
    \label{fig:triple_c}
  \end{subfigure}
  \caption{YOLOv11 animal detection on (a,d) ground truth, (b,e) hazy image, and (c,f) dehazed image}
  \label{fig:yolo_images}
\end{figure}

\subsection{Animal Detection}
To validate the efficacy of our approach for the downstream detection task, we train the YOLOv11\cite{yolov11} detection model on the NTLNP dataset and evaluate performance on hazy images from the AnimalHaze3k test set and the corresponding dehazed images generated by IncepdehazeGan. Performance was evaluated using mAP and IoU metrics.

    


Experimentation demonstrates the superiority of our approach at dehazing wildlife imagery. Achieving a remarkable SSIM of 0.8913 and PSNR of 20.54, our model surpasses existing SOTA methods (see \cref{tab:dehaze_comparison}). Our experiments on detection show a significant improvement in mAP ($>$112\%)  and mIoU ($>$67\%) metrics when using dehazed images generated by IncepDehazeGan (see \cref{tab:detection_results}). This demonstrates the effectiveness of our model at providing high-quality data for all computer-vision tasks related to wildlife monitoring (see \cref{fig:yolo_images}).

\begin{table}[h]
    \centering
    \captionsetup{justification=centering}
    \begin{tabular}{lcccc} 
       \toprule
       \textbf{Model Name} & \textbf{SSIM} $\uparrow$ & \textbf{PSNR} $\uparrow$ & \textbf{FSIM} $\uparrow$ & \textbf{LPIPS} $\downarrow$ \\
        \midrule
        FFA\cite{FFANET} & 0.5468 & 12.1842 & 0.6665 & 0.3610 \\
        FD GAN\cite{DBLP:journals/corr/abs-2001-06968} & 0.5580 & 17.5573 & 0.8314 & 0.1637 \\
        DEANET\cite{DEANET} & 0.8303 & 18.6481 & 0.8936 & 0.2102 \\
        DehazeFormer\cite{song2023vision} & 0.8388 & 17.4550 & 0.8917 & 0.2375 \\
        \textbf{IncepDehaze} & \textbf{0.8914} & \textbf{20.5404} & \textbf{0.9363} & \textbf{0.1104} \\
        \bottomrule
    \end{tabular}
    \caption{Performance of different dehazing methods on the AnimalHaze3k dataset.}
    \label{tab:dehaze_comparison}
\end{table}

\begin{table}[h]
    \centering
    \captionsetup{justification=centering}
    \begin{tabular}{lcc}
        \toprule
        \textbf{Images} & \textbf{mAP} $\uparrow$ & \textbf{mIoU} $\uparrow$ \\
        \midrule
        Hazy & 0.3216 & 0.4313 \\
        Dehazed Output & \textbf{0.6842} & \textbf{0.7201} \\
        \bottomrule
    \end{tabular}
    \caption{Animal detection performance of YOLOv11 on hazy images and dehazed counterparts generated IncepdeHazeGan.}
    \label{tab:detection_results}
\end{table}

\section{Conclusion}
\label{paper_sec:conclusion}

This work presents the AnimalHaze3k dataset, a synthetic single-image dehazing dataset addressing atmospheric degradation in wildlife imagery. For this purpose, we introduce the IncepDehazeGan model. Extensive experimental comparisons with current state-of-the-art models on various metrics (PSNR, SSIM, LPIPS and FSIM) confirm the effectiveness of our model in generating clearer images. Our experiment on the detection task demonstrates that IncepDehazeGan overcomes challenges posed by atmospheric haze and can provide higher quality data for tasks such as animal identification, population tracking, movement tracking, behavior pattern tracking and surveillance. Thus, it helps advance efforts for animal behavioral studies and conservation. 

{
    \small
    \bibliographystyle{ieeenat_fullname}
    \bibliography{paper.bib}

@article{hou2020identification,
  title={Identification of animal individuals using deep learning: A case study of giant panda},
  author={Hou, Jin and He, Yuxin and Yang, Hongbo and Connor, Thomas and Gao, Jie and Wang, Yujun and Zeng, Yichao and Zhang, Jindong and Huang, Jinyan and Zheng, Bochuan and others},
  journal={Biological Conservation},
  volume={242},
  pages={108414},
  year={2020},
  publisher={Elsevier}
}

@article{ren2021tracking,
  title={Tracking and analysing social interactions in dairy cattle with real-time locating system and machine learning},
  author={Ren, Keni and Bernes, Gun and Hetta, M{\aa}rten and Karlsson, Johannes},
  journal={Journal of Systems Architecture},
  volume={116},
  pages={102139},
  year={2021},
  publisher={Elsevier}
}

@article{guan2023face,
  title={Face recognition of a Lorisidae species based on computer vision},
  author={Guan, Yan and Lei, Yujie and Zhu, Yuhui and Li, Tingxuan and Xiang, Ying and Dong, Pengmei and Jiang, Rong and Luo, Jinwen and Huang, Anqi and Fan, Yumai and others},
  journal={Global Ecology and Conservation},
  volume={45},
  pages={e02511},
  year={2023},
  publisher={Elsevier}
}

@article{chalmers2019conservation,
  title={Conservation AI: Live stream analysis for the detection of endangered species using convolutional neural networks and drone technology},
  author={Chalmers, Carl and Fergus, Paul and Wich, Serge and Montanez, Aday Curbelo},
  journal={arXiv preprint arXiv:1910.07360},
  year={2019}
}

@article{norouzzadeh2018automatically,
  title={Automatically identifying, counting, and describing wild animals in camera-trap images with deep learning},
  author={Norouzzadeh, Mohammad Sadegh and Nguyen, Anh and Kosmala, Margaret and Swanson, Alexandra and Palmer, Meredith S and Packer, Craig and Clune, Jeff},
  journal={Proceedings of the National Academy of Sciences},
  volume={115},
  number={25},
  pages={E5716--E5725},
  year={2018},
  publisher={National Academy of Sciences}
}

@article{schindler2021identification,
  title={Identification of animals and recognition of their actions in wildlife videos using deep learning techniques},
  author={Schindler, Frank and Steinhage, Volker},
  journal={Ecological Informatics},
  volume={61},
  pages={101215},
  year={2021},
  publisher={Elsevier}
}

@article{saoud2025hubot,
  title={HuBot: A biomimicking mobile robot for non-disruptive bird behavior study},
  author={Saoud, Lyes Saad and Lesobre, Lo{\"\i}c and Sorato, Enrico and Al Qaydi, Saud and Hingrat, Yves and Seneviratne, Lakmal and Hussain, Irfan},
  journal={Ecological Informatics},
  volume={85},
  pages={102939},
  year={2025},
  publisher={Elsevier}
}

@book{o2011camera,
  title={Camera traps in animal ecology: methods and analyses},
  author={O'Connell, Allan F and Nichols, James D and Karanth, K Ullas},
  volume={271},
  year={2011},
  publisher={Springer}
}

@article{newey2015limitations,
  title={Limitations of recreational camera traps for wildlife management and conservation research: A practitioner’s perspective},
  author={Newey, Scott and Davidson, Paul and Nazir, Sajid and Fairhurst, Gorry and Verdicchio, Fabio and Irvine, R Justin and Van Der Wal, Ren{\'e}},
  journal={Ambio},
  volume={44},
  pages={624--635},
  year={2015},
  publisher={Springer}
}

@article{singh2019comprehensive,
  title={A comprehensive review of computational dehazing techniques},
  author={Singh, Dilbag and Kumar, Vijay},
  journal={Archives of Computational Methods in Engineering},
  volume={26},
  number={5},
  pages={1395--1413},
  year={2019},
  publisher={Springer}
}

@article{wang2017recent,
  title={Recent advances in image dehazing},
  author={Wang, Wencheng and Yuan, Xiaohui},
  journal={IEEE/CAA Journal of Automatica Sinica},
  volume={4},
  number={3},
  pages={410--436},
  year={2017},
  publisher={IEEE/CAA Journal of Automatica Sinica}
}

@article{narasimhan2002vision,
  title={Vision and the atmosphere},
  author={Narasimhan, Srinivasa G and Nayar, Shree K},
  journal={International journal of computer vision},
  volume={48},
  pages={233--254},
  year={2002},
  publisher={Springer}
}

@inproceedings{narasimhan2003interactive,
  title={Interactive (de) weathering of an image using physical models},
  author={Narasimhan, Srinivasa G and Nayar, Shree K},
  booktitle={IEEE Workshop on color and photometric Methods in computer Vision},
  volume={6},
  number={6.4},
  pages={1},
  year={2003},
  organization={France}
}

@article{cai2016dehazenet,
  title={Dehazenet: An end-to-end system for single image haze removal},
  author={Cai, Bolun and Xu, Xiangmin and Jia, Kui and Qing, Chunmei and Tao, Dacheng},
  journal={IEEE transactions on image processing},
  volume={25},
  number={11},
  pages={5187--5198},
  year={2016},
  publisher={IEEE}
}

@inproceedings{ren2016single,
  title={Single image dehazing via multi-scale convolutional neural networks},
  author={Ren, Wenqi and Liu, Si and Zhang, Hua and Pan, Jinshan and Cao, Xiaochun and Yang, Ming-Hsuan},
  booktitle={Computer Vision--ECCV 2016: 14th European Conference, Amsterdam, The Netherlands, October 11-14, 2016, Proceedings, Part II 14},
  pages={154--169},
  year={2016},
  organization={Springer}
}

@inproceedings{zhang2018densely,
  title={Densely connected pyramid dehazing network},
  author={Zhang, He and Patel, Vishal M},
  booktitle={Proceedings of the IEEE conference on computer vision and pattern recognition},
  pages={3194--3203},
  year={2018}
}

@inproceedings{dong2020physics,
  title={Physics-based feature dehazing networks},
  author={Dong, Jiangxin and Pan, Jinshan},
  booktitle={Computer Vision--ECCV 2020: 16th European Conference, Glasgow, UK, August 23--28, 2020, Proceedings, Part XXX 16},
  pages={188--204},
  year={2020},
  organization={Springer}
}

@inproceedings{dong2020multi,
  title={Multi-scale boosted dehazing network with dense feature fusion},
  author={Dong, Hang and Pan, Jinshan and Xiang, Lei and Hu, Zhe and Zhang, Xinyi and Wang, Fei and Yang, Ming-Hsuan},
  booktitle={Proceedings of the IEEE/CVF conference on computer vision and pattern recognition},
  pages={2157--2167},
  year={2020}
}

@inproceedings{wu2021contrastive,
  title={Contrastive learning for compact single image dehazing},
  author={Wu, Haiyan and Qu, Yanyun and Lin, Shaohui and Zhou, Jian and Qiao, Ruizhi and Zhang, Zhizhong and Xie, Yuan and Ma, Lizhuang},
  booktitle={Proceedings of the IEEE/CVF conference on computer vision and pattern recognition},
  pages={10551--10560},
  year={2021}
}

@article{vaswani2017attention,
  title={Attention is all you need},
  author={Vaswani, Ashish and Shazeer, Noam and Parmar, Niki and Uszkoreit, Jakob and Jones, Llion and Gomez, Aidan N and Kaiser, {\L}ukasz and Polosukhin, Illia},
  journal={Advances in neural information processing systems},
  volume={30},
  year={2017}
}

@article{song2023vision,
  title={Vision transformers for single image dehazing},
  author={Song, Yuda and He, Zhuqing and Qian, Hui and Du, Xin},
  journal={IEEE Transactions on Image Processing},
  volume={32},
  pages={1927--1941},
  year={2023},
  publisher={IEEE}
}

@article{tan2022animal,
  title={Animal detection and classification from camera trap images using different mainstream object detection architectures},
  author={Tan, Mengyu and Chao, Wentao and Cheng, Jo-Ku and Zhou, Mo and Ma, Yiwen and Jiang, Xinyi and Ge, Jianping and Yu, Lian and Feng, Limin},
  journal={Animals},
  volume={12},
  number={15},
  pages={1976},
  year={2022},
  publisher={MDPI}
}

@article{ganj2024hybriddepth,
  title={HybridDepth: Robust Metric Depth Fusion by Leveraging Depth from Focus and Single-Image Priors},
  author={Ganj, Ashkan and Su, Hang and Guo, Tian},
  journal={arXiv preprint arXiv:2407.18443},
  year={2024}
}

@article{bhat2023zoedepth,
  title={Zoedepth: Zero-shot transfer by combining relative and metric depth},
  author={Bhat, Shariq Farooq and Birkl, Reiner and Wofk, Diana and Wonka, Peter and M{\"u}ller, Matthias},
  journal={arXiv preprint arXiv:2302.12288},
  year={2023}
}

@inproceedings{yang2022deep,
  title={Deep depth from focus with differential focus volume},
  author={Yang, Fengting and Huang, Xiaolei and Zhou, Zihan},
  booktitle={Proceedings of the IEEE/CVF conference on computer vision and pattern recognition},
  pages={12642--12651},
  year={2022}
}

@inproceedings{yang2024depth,
  title={Depth anything: Unleashing the power of large-scale unlabeled data},
  author={Yang, Lihe and Kang, Bingyi and Huang, Zilong and Xu, Xiaogang and Feng, Jiashi and Zhao, Hengshuang},
  booktitle={Proceedings of the IEEE/CVF Conference on Computer Vision and Pattern Recognition},
  pages={10371--10381},
  year={2024}
}

@article{li2018benchmarking,
  title={Benchmarking single-image dehazing and beyond},
  author={Li, Boyi and Ren, Wenqi and Fu, Dengpan and Tao, Dacheng and Feng, Dan and Zeng, Wenjun and Wang, Zhangyang},
  journal={IEEE Transactions on Image Processing},
  volume={28},
  number={1},
  pages={492--505},
  year={2018},
  publisher={IEEE}
}

@article{DBLP:journals/corr/abs-2001-06968,
  author       = {Yu Dong and
                  Yihao Liu and
                  He Zhang and
                  Shifeng Chen and
                  Yu Qiao},
  title        = {{FD-GAN:} Generative Adversarial Networks with Fusion-discriminator
                  for Single Image Dehazing},
  journal      = {CoRR},
  volume       = {abs/2001.06968},
  year         = {2020},
  url          = {https://arxiv.org/abs/2001.06968},
  eprinttype    = {arXiv},
  eprint       = {2001.06968},
  bibsource    = {dblp computer science bibliography, https://dblp.org}
}

@article{FFANET,
  author       = {Xu Qin and
                  Zhilin Wang and
                  Yuanchao Bai and
                  Xiaodong Xie and
                  Huizhu Jia},
  title        = {FFA-Net: Feature Fusion Attention Network for Single Image Dehazing},
  journal      = {CoRR},
  volume       = {abs/1911.07559},
  year         = {2019},
  url          = {http://arxiv.org/abs/1911.07559},
  eprinttype    = {arXiv},
  eprint       = {1911.07559},
  bibsource    = {dblp computer science bibliography, https://dblp.org}
}

@misc{DEANET,
      title={DEA-Net: Single image dehazing based on detail-enhanced convolution and content-guided attention}, 
      author={Zixuan Chen and Zewei He and Zhe-Ming Lu},
      year={2023},
      eprint={2301.04805},
      archivePrefix={arXiv},
      primaryClass={cs.CV},
      url={https://arxiv.org/abs/2301.04805}, 
}

@misc{yolov11,
      title={YOLOv11: An Overview of the Key Architectural Enhancements}, 
      author={Rahima Khanam and Muhammad Hussain},
      year={2024},
      eprint={2410.17725},
      archivePrefix={arXiv},
      primaryClass={cs.CV},
      url={https://arxiv.org/abs/2410.17725}, 
}

@ARTICLE{ssim,
  author={Zhou Wang and Bovik, A.C. and Sheikh, H.R. and Simoncelli, E.P.},
  journal={IEEE Transactions on Image Processing}, 
  title={Image quality assessment: from error visibility to structural similarity}, 
  year={2004},
  volume={13},
  number={4},
  pages={600-612},
  doi={10.1109/TIP.2003.819861}}

@INPROCEEDINGS{psnr,
  author={Horé, Alain and Ziou, Djemel},
  booktitle={2010 20th International Conference on Pattern Recognition}, 
  title={Image Quality Metrics: PSNR vs. SSIM}, 
  year={2010},
  volume={},
  number={},
  pages={2366-2369},
  doi={10.1109/ICPR.2010.579}}

@ARTICLE{fsim,
  author={Zhang, Lin and Zhang, Lei and Mou, Xuanqin and Zhang, David},
  journal={IEEE Transactions on Image Processing}, 
  title={FSIM: A Feature Similarity Index for Image Quality Assessment}, 
  year={2011},
  volume={20},
  number={8},
  pages={2378-2386},
  doi={10.1109/TIP.2011.2109730}}

@article{lpips,
  author       = {Richard Zhang and
                  Phillip Isola and
                  Alexei A. Efros and
                  Eli Shechtman and
                  Oliver Wang},
  title        = {The Unreasonable Effectiveness of Deep Features as a Perceptual Metric},
  journal      = {CoRR},
  volume       = {abs/1801.03924},
  year         = {2018},
  url          = {http://arxiv.org/abs/1801.03924},
  eprinttype    = {arXiv},
  eprint       = {1801.03924},
  bibsource    = {dblp computer science bibliography, https://dblp.org}
}

@article{inception,
  author       = {Christian Szegedy and
                  Wei Liu and
                  Yangqing Jia and
                  Pierre Sermanet and
                  Scott E. Reed and
                  Dragomir Anguelov and
                  Dumitru Erhan and
                  Vincent Vanhoucke and
                  Andrew Rabinovich},
  title        = {Going Deeper with Convolutions},
  journal      = {CoRR},
  volume       = {abs/1409.4842},
  year         = {2014},
  url          = {http://arxiv.org/abs/1409.4842},
  eprinttype    = {arXiv},
  eprint       = {1409.4842},
  bibsource    = {dblp computer science bibliography, https://dblp.org}
}

@article{residuals,
  title={Dehazing network: Asymmetric unet based on physical model},
  author={Du, Yang and Li, Jun and Sheng, Qinghong and Zhu, Yuxin and Wang, Bo and Ling, Xiao},
  journal={IEEE Transactions on Geoscience and Remote Sensing},
  volume={62},
  pages={1--12},
  year={2024},
  publisher={IEEE}
}

@ARTICLE{multilayer,
  author={Ma, Chenhui and Mu, Xiaodong and Sha, Dexuan},
  journal={IEEE Access}, 
  title={Multi-Layers Feature Fusion of Convolutional Neural Network for Scene Classification of Remote Sensing}, 
  year={2019},
  volume={7},
  number={},
  pages={121685-121694},
  doi={10.1109/ACCESS.2019.2936215}}

@inproceedings{pix2pix,
  title={Image-to-image translation with conditional adversarial networks},
  author={Isola, Phillip and Zhu, Jun-Yan and Zhou, Tinghui and Efros, Alexei A},
  booktitle={Proceedings of the IEEE conference on computer vision and pattern recognition},
  pages={1125--1134},
  year={2017}
}
}


\end{document}